# Emotion detection of social data: APIs comparative study

Bilal Abu-Salih[1], Mohammad Alhabashneh[2], Dengya Zhu[2], Albara Awajan[3],

Yazan Alshamaileh[1], Bashar Al-Shboul[1], Mohammad Alshraideh[1]

[1]The University of Jordan, Amman, Jordan

[2]Curtin University, Perth Australia

[3]Al-Balqa Applied University

**Abstract:** The development of emotion detection technology has emerged as a highly valuable possibility in the corporate sector due to the nearly limitless uses of this new discipline, particularly with the unceasing propagation of social data. In recent years, the electronic marketplace has witnessed the establishment of a large number of start-up businesses with an almost sole focus on building new commercial and open-source tools and APIs for emotion detection and recognition. Yet, these tools and APIs must be continuously reviewed and evaluated, and their performances should be reported and discussed. There is a lack of research to empirically compare current emotion detection technologies in terms of the results obtained from each model using the same textual dataset. Also, there is a lack of comparative studies that apply benchmark comparison to social data. This study compares eight technologies: IBM Watson NLU, ParallelDots, Symanto – Ekman, Crystalfeel, Text to Emotion, Senpy, Textprobe, and NLP Cloud. The comparison was undertaken using two different datasets. The emotions from the chosen datasets were then derived using the incorporated APIs. The performance of these APIs was assessed using the aggregated scores that they delivered as well as the theoretically proven evaluation metrics such as the micro-average of accuracy, classification error, precision, recall, and f1-score. Lastly, the assessment of these APIs incorporating the evaluation measures is reported and discussed.

## 1. Introduction

The global emotion detection and recognition market is growing at a significant rate, thanks to the internet of things, wearable technology, social media data, and the explosive development of smartphone usage. With a Compound Annual Growth Rate (CAGR) of 18.7% from 2021 to 2030, the emotion detection and recognition market, which was valued at $18.8 billion in 2020, is anticipated to expand to $103.1 billion by 2030 (Md Shadaab Khan 2022). As investment continues to rise globally in response to the demand for emotion detection systems, businesses continue to implement advanced social media analytics to better understand the voice of their customers (Al-Okaily and Al-Okaily 2022; Al-Okaily, Alghazzawi, et al. 2022; Al-Okaily, Alqudah, et al. 2022; Al-Okaily 2021).

In this paper, we focus on emotion detection technologies and their application to the written text obtained from social data. The application of textual emotion detection spans from education (by mining students' emotions toward the learning process) to marketing (by listening to the voice of customers), to several other use cases (Sailunaz et al. 2018). Although some people might prefer to use audio and video to convey their opinions and emotions on social media, written text is still dominating and used as a core means for delivering content to social audiences (Shawabkeh et al. 2021). Therefore, removing ambiguity and extracting the factual meaning behind the textual content is not a trivial task (Wongthontham and Abu-Salih 2018; Abu-Salih et al. 2020). Various tech companies compete to deliver solutions and APIs that are claimed to provide the best understanding of the social emotions extracted from the social data. Hence, there is an ongoing need to conduct benchmark comparisons between these APIs.

Most of the current emotion detection comparisons have been carried out to state-of-the-art benchmark approaches in terms of incorporated methodology, embedded technology, experimental datasets, and evaluation metrics (Kim and Klinger 2018; Saganowski et al. 2020; Zad et al. 2021). Other attempts evaluated commercial and open-source tools and APIs in terms of mechanism, ease of use, availability, etc. (Garcia-Garcia, Penichet, and Lozano 2017). However, there is a lack of research to empirically compare current emotion detection technologies in terms of the results obtained from each model using the same textual dataset. Also, there is a lack of comparative studies that apply benchmark comparison to social data. This is imperative as such comparisons guide companies investing in these technologies to select the best-evaluated tools, specifically that such commercial tools and APIs continuously enhance their internal algorithms. Hence, recent methodological comparisons must always be undergone.

This study compares eight technologies, namely IBM Watson NLU, ParallelDots, Symanto – Ekman, Crystalfeel, Text to Emotion, Senpy, Textprobe, and NLP Cloud. The comparison was undertaken using two different datasets: (1) Emotions dataset for NLP that uses a graph-based technique to manage an annotated tweet corpus in order to create contextualised, pattern-based emotion characteristics. (2) Twitter Conversations dataset: which is a set of 300 tweets (replies) captured from Twitter conversations of five official Twitter accounts of certain Australian banks. This dataset is then manually annotated, by two of the authors, with the relevant emotion(s). Then, the incorporated APIs were used to extract the emotions from the selected datasets and evaluated using their resultant aggregated scores as well as theoretically proven evaluation metrics, including micro-average of accuracy, classification error, precision, recall, and f1-score. Finally, the assessment of these APIs incorporating the evaluation measures is reported and discussed. The experimental results indicate an apparent variance between the extracted emotions of each API in each dataset. This discrepancy between the extracted emotions for each API emphasises the importance of conducting this study, thereby obtaining a better view of the reliability of these APIs to tackle the designated task.

The remaining of this paper is organised as follows: Section 2 introduces the notion of emotion detection and describes the incorporated APIs and tools. Section 3 explains the incorporated methodology regarding data selection and evaluation measures. The experimental results are reported and discussed in Section 4. Section 5 concludes the paper.

## 2. Emotion detection: A definition and incorporated APIs

### 2.1 Emotion detection

Defining and interpreting the term *emotion* is challenging, not only due to the lack of a scientific consensus on the definition but also due to the sophisticated nature of the term that makes it a *notorious* problem ([Scherer 2005](#)). Therefore, various theoretical studies have perceived *emotion* from different perspectives and portrayed it using different elements of affect and feeling. Amongst such divergent analyses, three key models are commonly incorporated, namely Ekman's theory of basic emotions ([Ekman 1992](#)), Plutchik's wheel of emotion ([Plutchik and Kellerman 2013](#)), and Russel's circumplex model ([Russell 1980](#)). Further, other emotion modelling frameworks were also proposed in the literature such as Shaver ([Shaver et al. 1987](#)), Oatley ([Cambria et al. 2013](#)), OCC ([Ortony, Clore, and Collins 1990](#)), VAD ([Verma and Tiwary 2017](#)), and Lovheim ([Lövheim 2012](#)). The reader can refer to ([Kim and Klinger 2018](#)) and ([Alqahtani and Alothaim 2022](#)) to obtain further information on these models.

In computer science, this term is interchangeably used in an interdisciplinary affective computing domain that deals with designing systems that can process, analyse, and detect human emotions. Hence, emotion detection (a.k.a. emotion recognition) becomes a nascent NLP task that aims to infer a specific feeling(s) from different multimodal datasets incorporating sophisticated machine and learning algorithms. Emotions - such as *joy, sadness, anger, frustration, love, fear*, and other feelings can be recognised from various modalities including facial expressions, textual contents, spoken expression ([Zhang, Wan, and Ming 2015](#)) and even physiology that is measured from wearable devices ([Saganowski et al. 2020](#)).

In fact, emotion recognition and sentiment analysis are interchangeably being used sometimes to indicate the polarity of a user's attitude expressed by the *positive, negative,* or *neutral* feelings deduced from the underlying opinion ([Abu-Salih et al. 2022](#); [Abu-Salih 2021](#)). However, these two notions differ in their definitions. Oxford dictionary defines emotion as: "a strong feeling deriving from one's circumstances, mood, or relationships with others", whereas sentiment is defined as "a view or opinion that is held or expressed". Distinguishing these two terms is essential to properly use each term in its precise context and use the appropriate technology to tackle them. This paper focuses on the notion of emotion detection and relevant APIs and tools that offer designated technical solutions.

### 2.2 Emotion detection APIs

This section provides a brief discussion on eight open-source and commercial emotion detection platforms and services used in this study. These tools have been selected from two commonly used API marketplaces, namely RapidAPI[1] and APILayer[2].

**IBM Watson NLU™:** IBM Watson offers an ecosystem of an interconnected set of cognitive services that deliver a range of capabilities. IBM Watson presents robust multilingual technical solutions to address problems spanning from (un/semi)structured data analytics to industrial-specific contexts ([Gliozzo et al. 2017](#); [High 2012](#)). The IBM Watson NLU is a RESTful API service that detects emotions from the written text using linguistic analytics. This service allows the user to send the request through plain text, a JSON file, instant messages, voice transcripts, or an HTML document and return a JSON file containing the resultant emotions and their respective likelihood values. The produced emotions are *anger*, *disgust*, *fear*, *joy*, and *sadness*. Figure 1 illustrates the call flow to this service.

---

[1] https://rapidapi.com/

[2] https://apilayer.com/

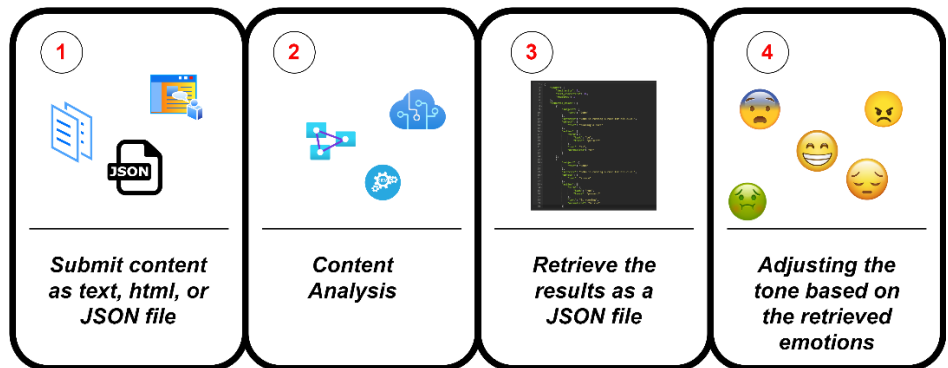

*Figure 1: IBM Watson™ Tone Analyzer call flow (IBM 2022)*

**ParallelDots API™:** ParallelDots[3] provides a paradigm of cognitive solutions built upon deep learning technology. The Text Analysis API supplied by ParallelDots incorporates other NLP techniques and is trained on over one billion documents. The services provided by ParallelDots include sentiment analysis, emotion detection, keyword extraction, named entity recognition, text classification, semantic analysis, etc. The ParallelDots' emotion detection service provides the capacity to analyse textual contents written in 15 languages and detect six emotion categories: *happy, angry, excited, sad, fear,* and *bored*.

**Symanto ™ – Emotion Text Analysis API:** Symanto platform[4] provides AI-powered tools that generate qualitative insights into the textual data. These insights include brand recommendation features, topic classification, psychographics analysis, timeline analysis, sentiment analysis, and emotion detection. As for Symanto's emotion analysis, Paul Ekman's Universal Emotions model is incorporated and used in the detection model. Hence, several emotion categories are inferred from the text as illustrated in Figure 2: *anger*, *contempt*, *disgust*, *enjoyment*, *fear*, *sadness*, and *surprise*.

**Text2Emotion – python library:** The Text2Emotion[5] library in Python allows the user to extract embedded emotions from a text in the form of a dictionary. The dictionary contains emotion categories as well as the emotion scores. The API is able to extract five different emotion emotion categories, namely *happy, angry, sad, surprise and fear*.

**Senpy™ Emotion Analysis:** Senpy[6] is a framework that is designed to build sentiment and emotion analysis services. It delivers functionalities for requesting sentiment and emotion detection from different sources using the same interface (API and vocabularies). The framework incorporates Ekkman and VAD emotion models and

---

[3] https://www.paralleldots.com/
[4] https://www.symanto.com/
[5] https://pypi.org/project/text2emotion/
[6] https://senpy.readthedocs.io/en/latest/examples.html

is built on evaluating algorithms with well-known datasets. The resultant emotions of a given text belong to one or more of the following emotions: f*ear, amusement, anger, annoyance, indifference, joy, awe, and sadness*,

**CrystalFeel™:** CrystalFeel[7] is a set of emotion analysis algorithms designed based on machine learning techniques for assessing emotional content in natural language. Grounded on a multi-theoretic conceptual foundation in emotion type, emotion dimension, and emotion intensity, CrystalFeel generates many psychologically relevant analytic outputs. In particular, CrysalFeel analyses emotional information in a text by concurrently executing five independently trained algorithms and reporting findings in five dimensions: *fear intensity, anger intensity, joy intensity, sadness intensity, and valence intensity*. It was created by A*STAR's Institute of High Performance Computing researchers who are exploring emotional and social intelligence[8].

**Textprobe™:** TextProbe[9] is a one-stop-shop text analysis API that automatically pulls a variety of insights from textual data, including the overall Sentiment (positive vs. negative) and emotional tone (joy, anger, fear, and sadness) The kernel of the API is built upon Ktrain[10] which is a lightweight wrapper for the TensorFlow Keras that makes it easier to construct, train, and deploy neural networks and other ML models.

**NLP Cloud™:** NLP Cloud[11] provides fast access endpoint for various NLP APIs including sentiment analysis, emotion detection, semantic similarity, text summarisation, etc. The emotion detection API is built upon DistilBERT base model (uncased)[12], a distilled version of the model proposed in (Sanh et al. 2019) which has been fine-tuned to infer emotions such as *love, joy, sadness, anger, fear, and surprise* in textual data.

Table 1 shows a summary of the selected emotion detection APIs and the emotion categories that are detected using each designated API.

*Table 1: The emotion categories extracted by each selected API.*

| Emotion | IBM Watson NLU | ParallelDots | Symanto – Ekman | Crystalfeel | Text to Emotion | Senpy | Textprobe | NLP Cloud |
|---|---|---|---|---|---|---|---|---|
| **Anger** | √ | √ | √ | √ | √ | √ | √ | √ |
| **Fear** | √ | √ | √ | √ | √ | √ | √ | √ |
| **Joy** | √ | √ (Happy) | √ | √ | √ (Happy) | √ | √ | √ |
| **Sadness** | √ | √ | √ | √ | √ | √ | √ | √ |
| **Exciting** | | √ | | | √ (surprise) | √ (awe) | | √ (surprise) |
| **Analytical** | √ | | | | | | | |
| **Valence** | | | | √ | | | | |
| **Indifference** | | | | | | √ | | |
| **Amusement** | | | | | | √ | | |
| **Bored** | | √ | | | | | | |
| **Love** | | | | | | | | √ |

---

[7] https://socialanalyticsplus.net/crystalfeel/
[8] https://www.a-star.edu.sg/ihpc
[9] https://textprobe.com/
[10] https://github.com/amaiya/ktrain
[11] https://nlpcloud.io/
[12] https://huggingface.co/distilbert-base-uncased

The conducted comparison in this study will focus on the four most common emotions amongst the selected APIs, namely *anger, fear, joy, and sadness*.

# 3. Method

This section discusses the proposed comparison methodology. First, we discuss and describe the datasets used in the comparison along with the technique used in preprocessing. Then, statistics on the extracted emotions using all APIs over both datasets are provided and discussed. This is followed by describing the incorporated evaluation measures. The section concludes with the experimental results.

## 3.1 Datasets selection and preprocessing

This study incorporates two different datasets:

**Dataset1 - Emotions dataset for NLP[13]:** This is an annotated tweet corpus managed by means of a graph-based technique which was used in the construction of contextualised, pattern-based emotion features (Sanh et al. 2019). These features were enhanced by word embeddings to retain the semantic relationships between patterns. The performance of the patterns was evaluated using several machine learning classifiers. This dataset contains 20,000 texts, and each is annotated using one of the following six emotions: *angry, fear, joy, sad, surprise, and love*. Figure 3 depicts the distribution of emotions dataset for NLP (Dataset1).

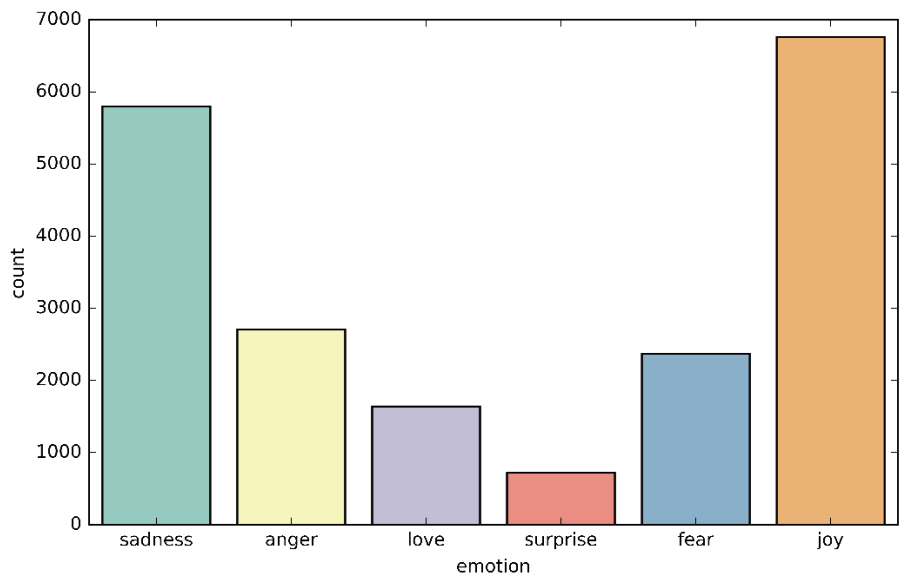

*Figure 3: Emotions distribution of Dataset1.*

**Dataset2 -Twitter Conversations dataset:** We collected around 300 tweets (replies) captured from Twitter conversations of five Twitter accounts which are the official Twitter accounts of certain Australian banks. Replies to tweets that are posted by service providers commonly covey customers' opinions toward provided service (Abu-Salih et al. 2021; Abu-Salih, Wongthongtham, and Kit 2018). Thus, these tweets would carry a rich source of emotions to be analysed. Also, we assembled this dataset to offer an unbiased benchmark comparison. This subjectively selected dataset has not been used to train any of the nominated APIs. Then, a labelling process was conducted by two of the authors to annotate each tweet with the relevant emotion(s). To achieve reliable data annotation, pairwise agreement and inter-coder reliability were computed based on the percentage of the total pairwise comparisons between the labellers. With an average of 97.53 percent, the overall inter-coder agreement rate varied from 93.45 to 100.00 percent. These rates are significantly higher than the minimum recommended rate of 70 percent. The inter-code agreement that was obtained as a result implies that the labelling process was extremely reliable and repeatable. Figure 4 shows the distribution of the resultant emotions captured from Dataset2. Online social networks have proven utility to provide an open environment for customers to convey their annoyance and frustration (Swaminathan and Mah 2016). This can be observed in the high proportion of *angry* emotions as depicted in Figure 4.

---

[13] https://www.kaggle.com/praveengovi/emotions-dataset-for-nlp

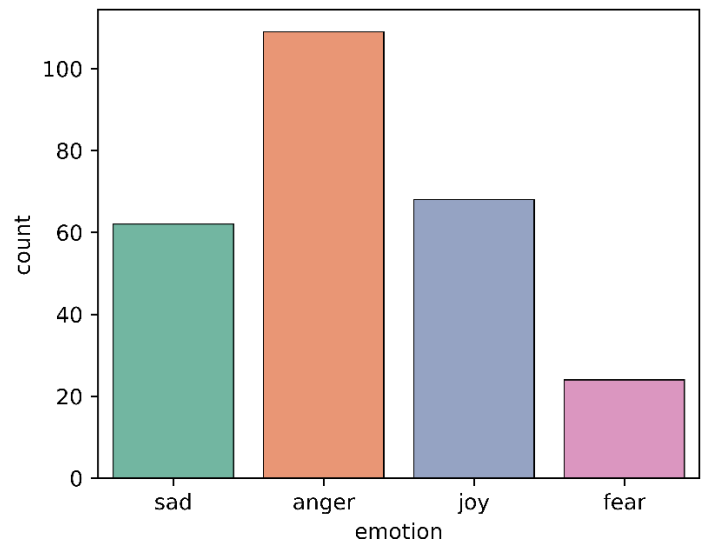

*Figure 4: Emotions distribution of Dataset2.*

**Data Cleansing:** The data cleansing process was undertaken to remove errors and nonsensical data. Also, we eliminate media contents such as images shared on Twitter or uploaded to one of the media sharing websites listed in (Saravanakumar and SuganthaLakshmi 2012) such as Instagram, Flickr, YouTube, and Pinterest. This is due to the fact that there is no text that can be extracted and used for analysis.

## 3.2 Emotions extraction

Each of the designated APIs listed in Section 2.2 was accessed to extract the emotions and their likelihood values (0.0 to 1.0) from the textual snippets of the two datasets discussed in the previous section. This was carried out using Python script, and the extracted emotions and their likelihood values of each API are stored in the MySQL database. Figure 5 and Figure 6 depict the aggregated count of retrieved emotion scores for each API applied on Dataset1 and Dataset2, respectively.

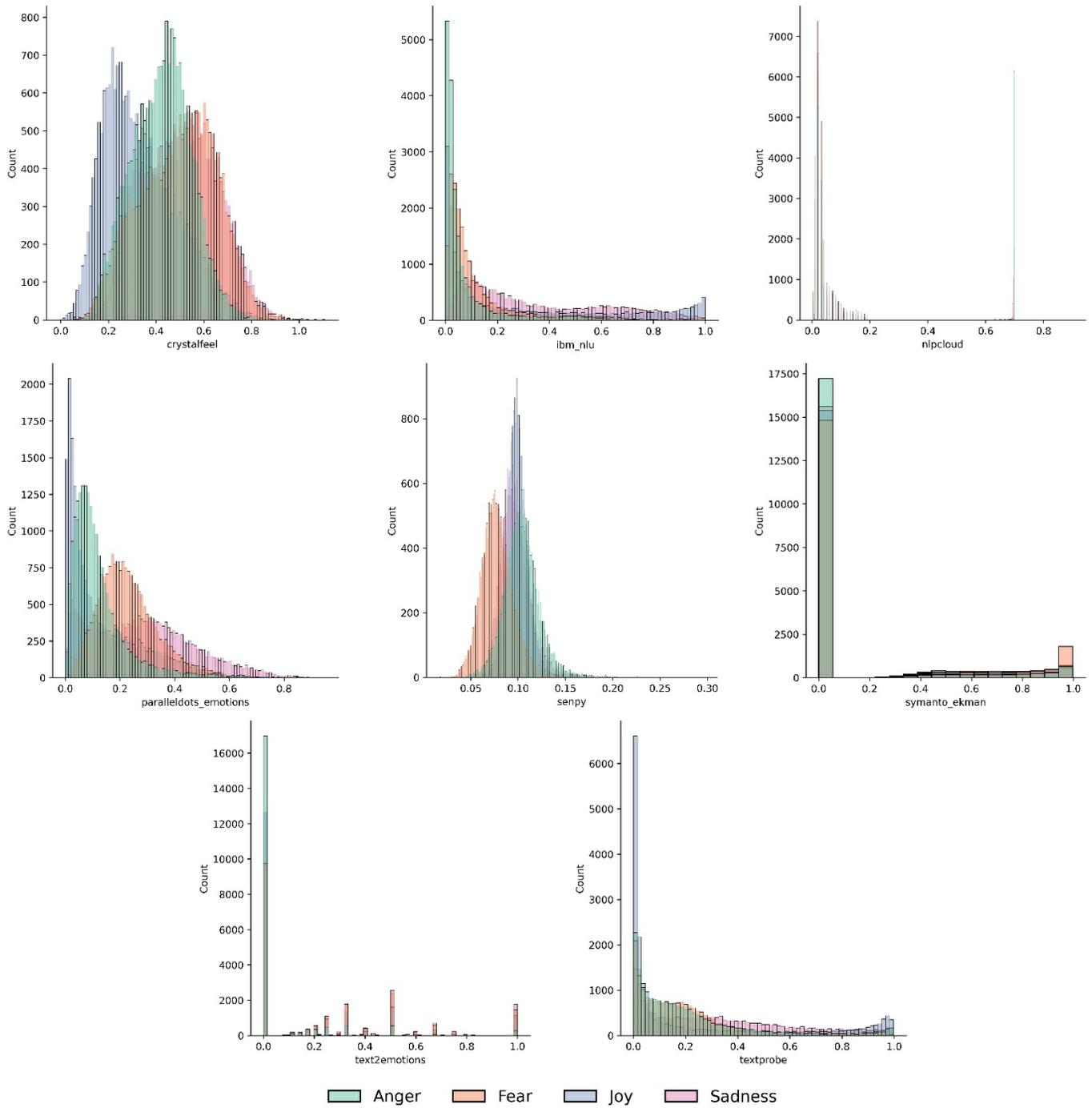

*Figure 5: The aggregated count of retrieved emotion scores for each API applied on Dataset1*

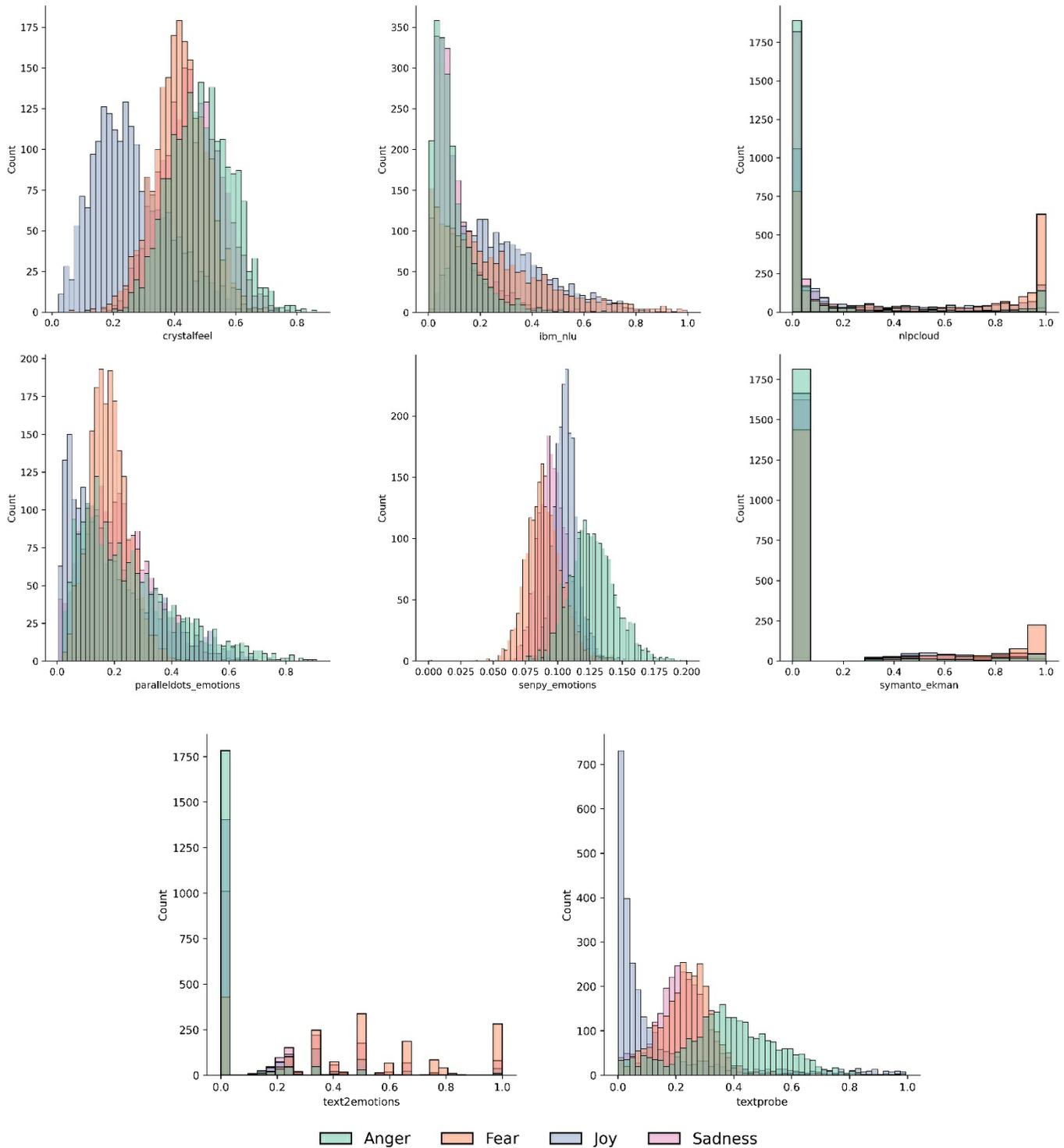

*Figure 6: The aggregated count of retrieved emotion scores for each API applied on Dataset2*

**Correlation Analysis:** To provide a further insight into the interrelation between the emotion scores obtained by each API, a correlation analysis is undertaken. Figure 7 and Figure 8 depict the correlation of emotion values obtained by each designated API when applying on Dataset1 and Dataset2, respectively. Figure 7 and Figure 8 verify again the weak and sometimes the negative correlation between various APIs in four emotions. However, there are mild and strong correlations between some APIs. For example, there is a relatively high correlation between paralleldots, ibm_nlu, and crystalfeel APIs in detecting *anger and joy* emotions captured from Dataset1(Figure 7), and there is a high correlation between paralleldots, ibm_nlu, textprobe and crystalfeel in detecting *joy* emotion captured from Dataset2 (Figure 8). In general, the poor correlation between several APIs requires conducting further scrutiny on these APIs to measure their performance. Next section discusses the proposed comparison approach.

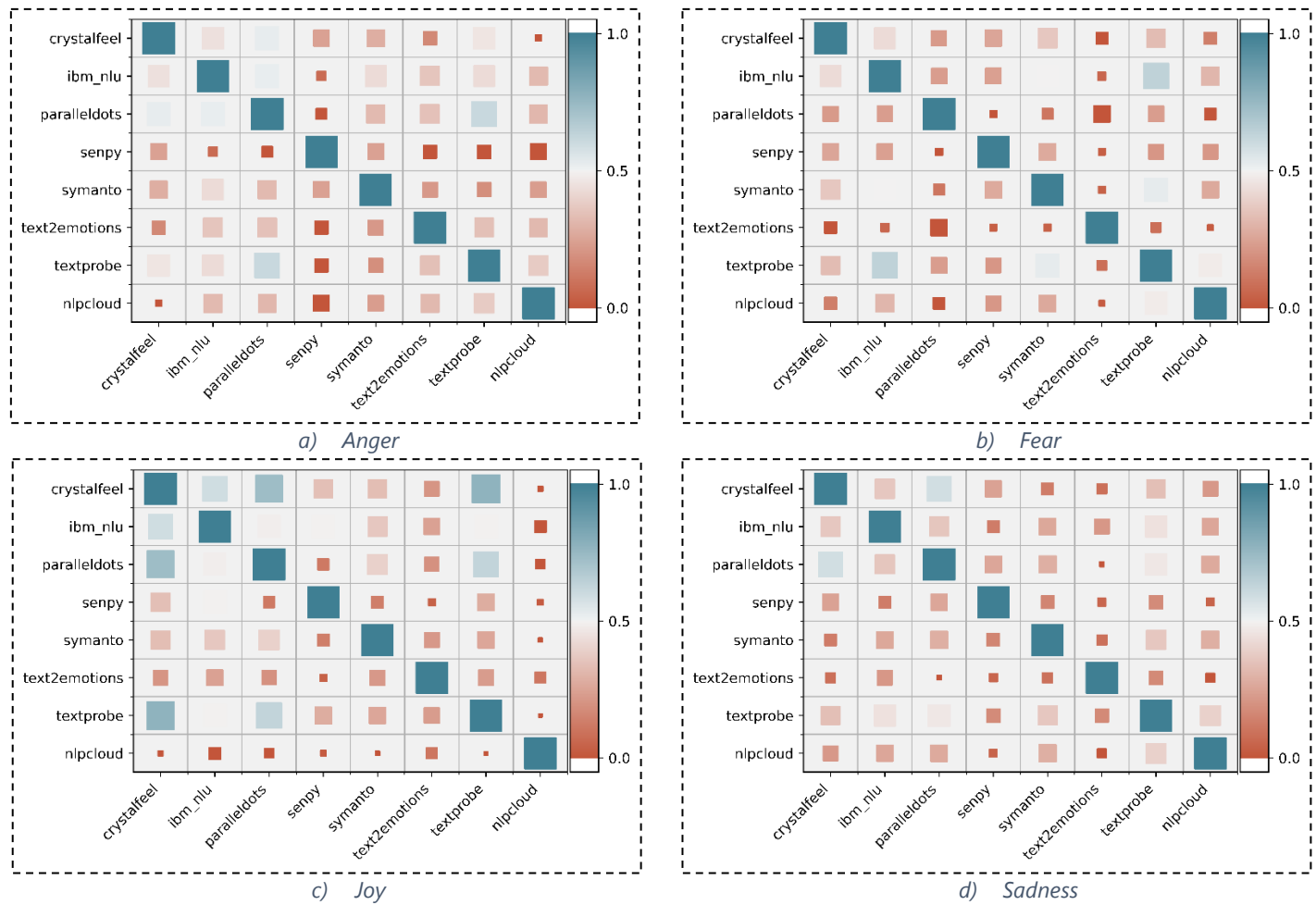

*Figure 7: The Correlation between different APIs in each inferred emotions captured from Dataset1.*

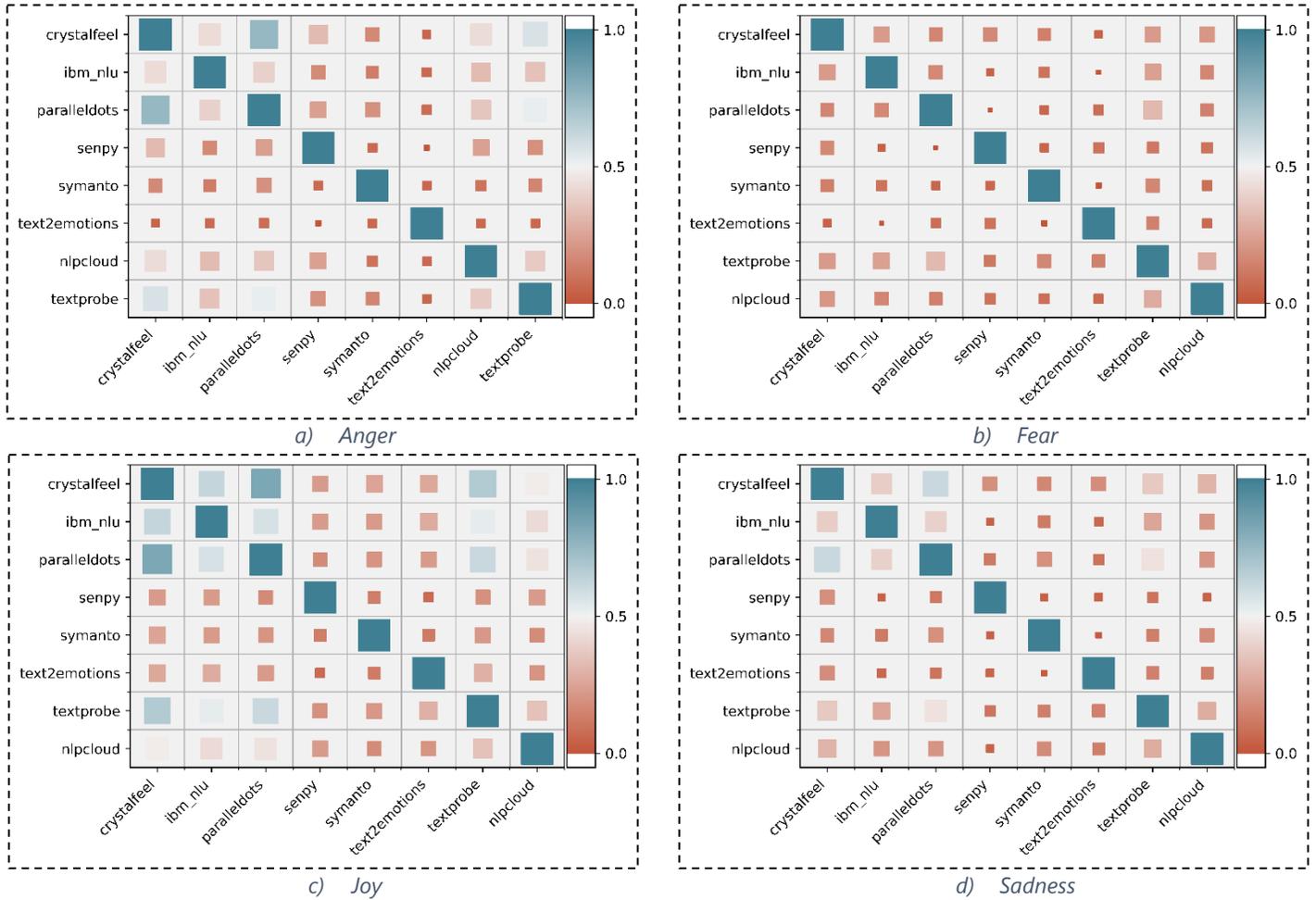

Figure 8: The Correlation between different APIs in inferred emotions captured from Dataset2.

## 3.3 Evaluation Measures

The aim of this experiment is to test each model's capacity to classify the textual contents to the designated emotions truly. We incorporate two evaluation mechanisms to measure the performance of the selected API:

**1) Evaluation metrics:** To evaluate the performance of these APIs, we define the essential structure blocks of the incorporated evaluation measures, namely True Positive (TP), False Positive (FP), False Negative (FN), and True Negative (TN). Figure 9 demonstrates these four scenarios in detecting whether a textual snippet depicts an *anger* emotion or not. However, these scenarios are used in all emotions so as to build the evaluation measures. The four scenarios in detecting *anger* emotion can be described as follows:

- TP: is where the API correctly detects an anger emotion as it appears in the designated dataset (correct classification).
- FN: is where the API detects a non-anger emotion of a text that is labelled as anger in the designated dataset (incorrect classification).
- FP: is where the API detects an anger emotion of a text that is labelled as non-anger in the designated dataset (incorrect classification).

- TN: is where the API correctly detects a non-anger emotion as it appears in the designated dataset (correct classification).

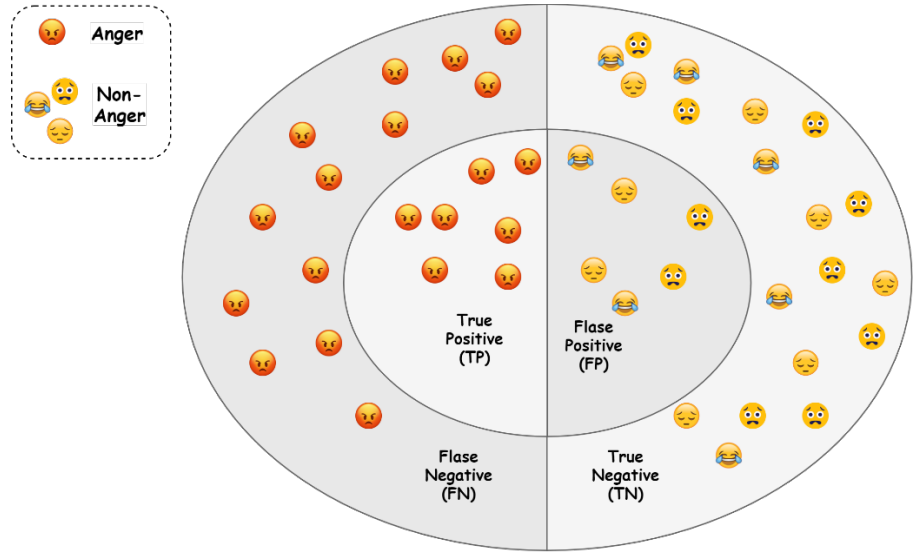

*Figure 9: Four scenarios in the API results in detecting angry emotion.*

These emotion detection scenarios are used in various evaluation metrics that are incorporated to validate the utility of the APIs over the two datasets. To provide an aggregated evaluation mechanism, micro average is computed for Classification error, Accuracy, Precision, Recall, and F1 as follows:

**(i)** **Micro-Accuracy**: Accuracy indicates the ability of the API to carry out correct emotion detections, which can be represented by the ratio between the actual number of accurate detections (i.e., TP + TN) to the total number of detections (FN + TP + FP + TN). Micro-accuracy indicates the average accuracy of each API to detect each emotion correctly, and it is computed as follows:

$$Micro - Accuracy = \frac{\sum_{i=1}^{n} TP^i + \sum_{i=1}^{n} TN^i}{\sum_{i=1}^{n} TP^i + \sum_{i=1}^{n} FP^i + \sum_{i=1}^{n} FN^i + \sum_{i=1}^{n} TN^i} \quad (1)$$

**(ii)** **Micro-Classification error:** Classification error denotes the ratio between the total number of incorrect detection (FP + FN) to the total number of detections (FN + TP + FP + TN). Micro-classification error is the average classification error which can be formulated as:

$$Micro - Classification\ error = \frac{\sum_{i=1}^{n} FP^i + \sum_{i=1}^{n} FN^i}{\sum_{i=1}^{n} TP^i + \sum_{i=1}^{n} FP^i + \sum_{i=1}^{n} FN^i + \sum_{i=1}^{n} TN^i} \quad (2)$$

**(iii)** **Micro-Precision:** Precision metric refers to the proportion of the number of textual snippets that were accurately detected to the total number of correct and incorrect detections based on the labelled dataset. Micro-Precision measures the average precision obtained by all emotions. It can be calculated as:

$$Micro - Precision = \frac{\sum_{i=1}^{n} TP^i}{\sum_{i=1}^{n} TP^i + \sum_{i=1}^{n} FP^i} \quad (3)$$

(iv) **Micro-Recall:** Recall refers to the ratio between the number of textual snippets that were accurately detected to the total number of actual emotions as per the labelled dataset. The micro recall represents the global average recall score. It can be computed as:

$$Micro - Recall = \frac{\sum_{i=1}^{n} TP^i}{\sum_{i=1}^{n} TP^i + \sum_{i=1}^{n} FN^i} \qquad (4)$$

(v) **Micro-F1:** F1 score (a.k.a. F-measure) represents a trade-off between the values obtained by both precision and recall. Micro-F1 denotes the global average F1-score which can be computed as:

$$Micro - F1 = 2 \times \frac{Micro-precision \times Micro-recall}{Micro-precision + Micro-recall} \qquad (5)$$

In the previous formulas, $n$ represents the number of emotions (i.e., 4 in our study). We compute the classification scenarios (i.e., TP, FN, FP, and TN) for all emotions obtained by all APIs, and then the average value is computed for each evaluation metric as depicted in the above equations.

In addition, **macro-average** of each of these metrics over the obtained emotions is also computed. Macro-average indicates the mean of each obtained metric per emotion which signifies a balancing metric that treat all classes equally. The generic equation of macro-average is defined as follows:

$$Macro - metric = \frac{\sum_{i=1}^{n} metric^i}{n} \qquad (6)$$

Where $metric$ demotes to one of the five incorporated metrics, namely Classification error, Accuracy, Precision, Recall, and F1, and $n$ indicates the number of emotions. Equation 6 indicates the mechanism followed to calculate the macro-average of each metric for a certain API. For example, to find the macro-average of Precision, we sum the Precision values obtained by each emotion then divided the sum over the number of emotions we have (i.e., 4 in our study).

**2) Comparison between the aggregated scores:** In this test, we aggregate all the emotional likelihood values obtained by each API on each dataset. Then the average of these values is computed and compared.

## 4 Results

### 4.1 Comparison results based on the evaluation metrics

In this experiment, the four detection scenarios (i.e., TP, FN, FP, and TN) were recorded for all APIs based on their performances on all extracted emotions. These values will be used to compute both the micro-average as well as the macro-average of each metric per API. As for the micro-average, the summation of each of the detection scenarios are computed (i.e., $TP_{sum}$, $FN_{sum}$, $FP_{sum}$, and $TN_{sum}$). This is to prepare data for calculating the micro average of classification error, accuracy, precision, recall, and f1-score as illustrated in equations (1) to (5). Table 2 and Table 3 demonstrate the micro-average values of these metrics based on the experiments conducted on both dataset1 and dataset2 respectively.

Table 2: Micro-average of classification error, accuracy, precision, recall, and f1-score based on dataset1.

| Model | TP$_{sum}$ ($\sum_{i=1}^{n} TP^i$) | FN$_{sum}$ ($\sum_{i=1}^{n} FN^i$) | FP$_{sum}$ ($\sum_{i=1}^{n} FP^i$) | TN$_{sum}$ ($\sum_{i=1}^{n} TN^i$) | Micro-Class. error (%) | Micro-Accuracy (%) | Micro-Precision (%) | Micro-Recall (%) | Micro-F1-score (%) |
|---|---|---|---|---|---|---|---|---|---|
| Crystalfeel | 11,025 | 6,615 | 16,156 | 54,404 | 25.82 | 74.18 | 40.56 | 62.50 | 49.20 |
| Ibm_nlu | 8,040 | 9,600 | 4,156 | 66,404 | 15.60 | 84.40 | 65.92 | 45.58 | 53.89 |
| Nlpcloud | 17,308 | 332 | 154 | 70,406 | 0.55 | 99.45 | 99.12 | 98.12 | 98.62 |
| Paralleldots | 2,376 | 15,264 | 1,162 | 69,398 | 18.62 | 81.38 | 67.16 | 13.47 | 22.44 |
| Senpy | 759 | 16,881 | 922 | 69,638 | 20.18 | 79.82 | 45.15 | 4.30 | 7.86 |
| Symanto_ekman | 6,069 | 11,571 | 6,163 | 64,397 | 20.11 | 79.89 | 49.62 | 34.40 | 40.63 |
| Text2emotions | 5,648 | 11,992 | 7,479 | 63,081 | 22.08 | 77.92 | 43.03 | 32.02 | 36.71 |
| Textprobe | 9,424 | 8,216 | 2,814 | 67,746 | 12.51 | 87.49 | 77.01 | 53.42 | 63.08 |

Table 3: Micro-average of classification error, accuracy, precision, recall, and f1-score based on dataset2.

| Model | TP$_{sum}$ ($\sum_{i=1}^{n} TP^i$) | FN$_{sum}$ ($\sum_{i=1}^{n} FN^i$) | FP$_{sum}$ ($\sum_{i=1}^{n} FP^i$) | TN$_{sum}$ ($\sum_{i=1}^{n} TN^i$) | Micro-Class. error (%) | Micro-Accuracy (%) | Micro-Precision (%) | Micro-Recall (%) | Micro-F1-score (%) |
|---|---|---|---|---|---|---|---|---|---|
| Crystalfeel | 212 | 109 | 149 | 751 | 21.13 | 78.87 | 58.73 | 66.04 | 62.17 |
| Ibm_nlu | 51 | 270 | 14 | 886 | 23.26 | 76.74 | 78.46 | 15.89 | 26.42 |
| Nlpcloud | 158 | 163 | 94 | 806 | 21.05 | 78.95 | 62.70 | 49.22 | 55.15 |
| Paralleldots | 75 | 246 | 13 | 887 | 21.21 | 78.79 | 85.23 | 23.36 | 36.67 |
| Senpy | 24 | 307 | 20 | 886 | 26.43 | 73.57 | 54.55 | 7.25 | 12.80 |
| Symanto_ekman | 39 | 282 | 86 | 814 | 30.14 | 69.86 | 31.20 | 12.15 | 17.49 |
| Text2emotions | 34 | 287 | 138 | 762 | 34.81 | 65.19 | 19.77 | 10.59 | 13.79 |
| Textprobe | 153 | 168 | 21 | 879 | 15.48 | 84.52 | 87.93 | 47.66 | 61.82 |

As depicted in Table 2, *nlpcloud* demonstrates superiority over all other APIs in detecting emotions using dataset1. This is because the micro-average of all metrics overshadows different APIs' values. *Textprope* API also shows good performance in terms of micro F1-score. On the contrary, *Senpy* API exhibits poor performance in this experiment in terms of the obtained micro F1-score. Table 3 shows the evaluation results of all APIs using dataset2. As discussed, this dataset has not been used previously and is the first paper that uses this dataset for emotion analysis. The aim of this paper is to conduct an unbiased assessment of the performance of each API using an unseen and untrained dataset. The experimental results using this dataset, as seen in Table 3, show moderate performances amongst most APIs. Although specific APIs such as *Crystalfeel* and *Textprobe* verify their utility to a certain extent, most other APIs failed to exhibit high performance in this experiment regarding the micro F1-score metric.

To compute the macro-average of each metric per API, the value of each metric per emotion is computed and then aggregated using equation 6. Table 4 and Table 5 demonstrate the macro-average values of these metrics based on the experiments conducted on both dataset1 and dataset2 respectively. Despite some slight improvements or deteriorations on the macro-average values obtained for certain APIs, the exhibited figures in these tables emphasise again that most of the incorporated APIs failed to tackle the datasets properly in terms of the results obtained from each model using the same textual dataset.

Table 4: Micro-average of classification error, accuracy, precision, recall, and f1-score based on dataset1

| Metric | Emotion | Crystalfeel (%) | ibm_nlu (%) | Nlpcloud (%) | Paralleldots (%) | Senpy (%) | symanto_ekman (%) | text2emotions (%) | Textprobe (%) |
|---|---|---|---|---|---|---|---|---|---|
| Classification error | Anger | 24.61 | 14.76 | 0.55 | 14.19 | 15.84 | 17.48 | 15.57 | 10.02 |
| | Fear | 40.63 | 11.73 | 0.57 | 13.76 | 14.19 | 20.27 | 27.92 | 9.32 |
| | Joy | 21.79 | 16.48 | 0.66 | 36.41 | 37.88 | 26.54 | 29.52 | 17.69 |
| | Sadness | 25.12 | 26.11 | 0.43 | 25.42 | 31.87 | 27.86 | 28.45 | 17.29 |
| Macro- Classification error | | **28.04** | **17.27** | **0.55** | **22.44** | **24.94** | **23.04** | **25.36** | **13.58** |
| Accuracy | Anger | 76.68 | 86.91 | 99.45 | 87.52 | 86.18 | 84.16 | 86.22 | 90.63 |
| | Fear | 60.14 | 89.23 | 99.43 | 87.76 | 87.39 | 80.82 | 74.03 | 91.23 |
| | Joy | 81.98 | 85.44 | 99.34 | 73.24 | 72.45 | 78.40 | 76.40 | 84.73 |
| | Sadness | 75.88 | 77.02 | 99.57 | 79.06 | 75.49 | 76.95 | 75.63 | 84.46 |
| Macro- Accuracy | | **73.67** | **84.65** | **99.45** | **81.90** | **80.38** | **80.08** | **78.07** | **87.76** |
| Precision | Anger | 30.83 | 53.08 | 98.22 | 75.60 | 34.80 | 37.88 | 44.30 | 70.58 |
| | Fear | 20.78 | 58.75 | 97.84 | 40.36 | 33.48 | 31.92 | 20.48 | 67.56 |
| | Joy | 93.72 | 84.54 | 99.77 | 81.99 | 61.40 | 75.25 | 68.38 | 89.82 |
| | Sadness | 50.75 | 53.71 | 99.31 | 68.14 | 53.73 | 56.35 | 50.90 | 72.84 |
| Macro- Precision | | **49.02** | **62.52** | **98.78** | **66.52** | **45.85** | **50.35** | **46.01** | **75.20** |
| Recall | Anger | 60.43 | 14.62 | 97.67 | 9.27 | 4.39 | 29.64 | 13.62 | 50.83 |
| | Fear | 83.99 | 30.85 | 97.35 | 6.70 | 6.45 | 54.49 | 41.30 | 50.11 |
| | Joy | 37.49 | 58.07 | 97.84 | 4.38 | 1.55 | 32.84 | 27.60 | 50.63 |
| | Sadness | 83.84 | 51.51 | 98.96 | 28.81 | 6.59 | 30.24 | 41.97 | 59.25 |
| Macro- Recall | | **66.44** | **38.76** | **97.96** | **12.29** | **4.75** | **36.80** | **31.12** | **52.70** |
| F-measure | Anger | 40.83 | 22.92 | 97.95 | 16.51 | 7.80 | 33.26 | 20.84 | 59.10 |
| | Fear | 33.32 | 40.45 | 97.59 | 11.49 | 10.81 | 40.26 | 27.38 | 57.54 |
| | Joy | 53.56 | 68.85 | 98.80 | 8.31 | 3.03 | 45.72 | 39.33 | 64.76 |
| | Sadness | 63.23 | 52.58 | 99.14 | 40.49 | 11.74 | 39.36 | 46.01 | 65.35 |
| Macro- F-measure | | **47.73** | **46.20** | **98.37** | **19.20** | **8.35** | **39.65** | **33.39** | **61.68** |

Table 5: Micro-average of classification error, accuracy, precision, recall, and f1-score based on dataset2

| Metric | Emotion | Crystalfeel (%) | ibm_nlu (%) | Nlpcloud (%) | Paralleldots (%) | Senpy (%) | symanto_ekman (%) | text2emotions (%) | Textprobe (%) |
|---|---|---|---|---|---|---|---|---|---|
| Classification error | Anger | 21.55 | 42.04 | 26.97 | 33.33 | 42.24 | 41.30 | 43.36 | 21.90 |
| | Fear | 26.60 | 42.04 | 23.22 | 30.49 | 40.09 | 57.83 | 87.61 | 17.88 |
| | Joy | 14.90 | 15.77 | 22.70 | 17.39 | 30.71 | 31.67 | 25.83 | 12.60 |
| | Sadness | 29.28 | 22.18 | 24.38 | 25.22 | 27.56 | 26.78 | 27.69 | 19.34 |
| Macro- Classification error | | **23.08** | **30.50** | **24.32** | **26.61** | **35.15** | **39.39** | **46.12** | **17.93** |
| Accuracy | Anger | 80.06 | 70.40 | 77.57 | 74.45 | 69.47 | 70.40 | 69.47 | 81.31 |
| | Fear | 75.39 | 89.23 | 99.43 | 87.76 | 87.39 | 80.82 | 74.03 | 91.23 |
| | Joy | 87.03 | 85.44 | 99.34 | 73.24 | 72.45 | 78.40 | 76.40 | 84.73 |
| | Sadness | 73.08 | 77.02 | 99.57 | 79.06 | 75.49 | 76.95 | 75.63 | 84.46 |
| Macro- Accuracy | | **78.89** | **77.07** | **78.95** | **78.95** | **73.72** | **70.16** | **65.80** | **84.59** |
| Precision | Anger | 64.29 | 100.00 | 70.00 | 75.00 | 43.75 | 55.56 | 25.00 | 79.03 |
| | Fear | 56.69 | 100.00 | 84.00 | 100.00 | 63.64 | 10.64 | 0.96 | 96.08 |
|  | Joy | 100.00 | 81.40 | 51.82 | 87.50 | 62.50 | 39.47 | 62.50 | 94.87 |
|  | Sadness | 41.30 | 70.00 | 53.13 | 71.43 | 0.00 | 45.16 | 42.50 | 81.82 |
| **Macro- Precision** |  | **65.57** | **87.85** | **64.74** | **83.48** | **42.47** | **37.71** | **32.74** | **87.95** |
| **Recall** | Anger | 75.00 | 1.04 | 43.75 | 21.88 | 7.29 | 5.21 | 1.04 | 51.04 |
|  | Fear | 75.00 | 1.04 | 43.75 | 21.88 | 7.29 | 5.21 | 1.04 | 51.04 |
|  | Joy | 44.12 | 51.47 | 83.82 | 41.18 | 12.82 | 22.06 | 22.06 | 54.41 |
|  | Sadness | 62.30 | 22.95 | 27.87 | 8.20 | 0.00 | 22.95 | 27.87 | 29.51 |
| **Macro- Recall** |  | **64.10** | **19.13** | **49.80** | **23.28** | **6.85** | **13.86** | **13.00** | **46.50** |
| **F-measure** | Anger | 69.23 | 2.06 | 53.85 | 33.87 | 12.50 | 9.52 | 2.00 | 62.03 |
|  | Fear | 64.57 | 2.06 | 57.53 | 35.90 | 13.08 | 6.99 | 1.00 | 66.67 |
|  | Joy | 61.22 | 63.06 | 64.04 | 56.00 | 21.28 | 28.30 | 32.61 | 69.16 |
|  | Sadness | 49.67 | 34.57 | 36.56 | 14.71 | 0 | 30.43 | 33.66 | 43.37 |
| **Macro- F-measure** |  | **61.18** | **25.44** | **53.00** | **35.12** | **0** | **18.81** | **17.32** | **60.31** |

## 4.2 Comparison results based on the aggregated scores

The obtained correlation between the APIs as discussed in Section 3.2 and illustrated in Figure 7 and Figure 8 shows an evident inconsistency between the performances of various APIs. To find the effectiveness of each API in extracting the emotions captured from each dataset, an accuracy comparison is undertaken. First, we aggregate all emotions likelihood values obtained by each API on each dataset. Then the average of these values is computed and compared. Figure 10 illustrates the average emotions likelihood values obtained by each API when applied to each dataset.

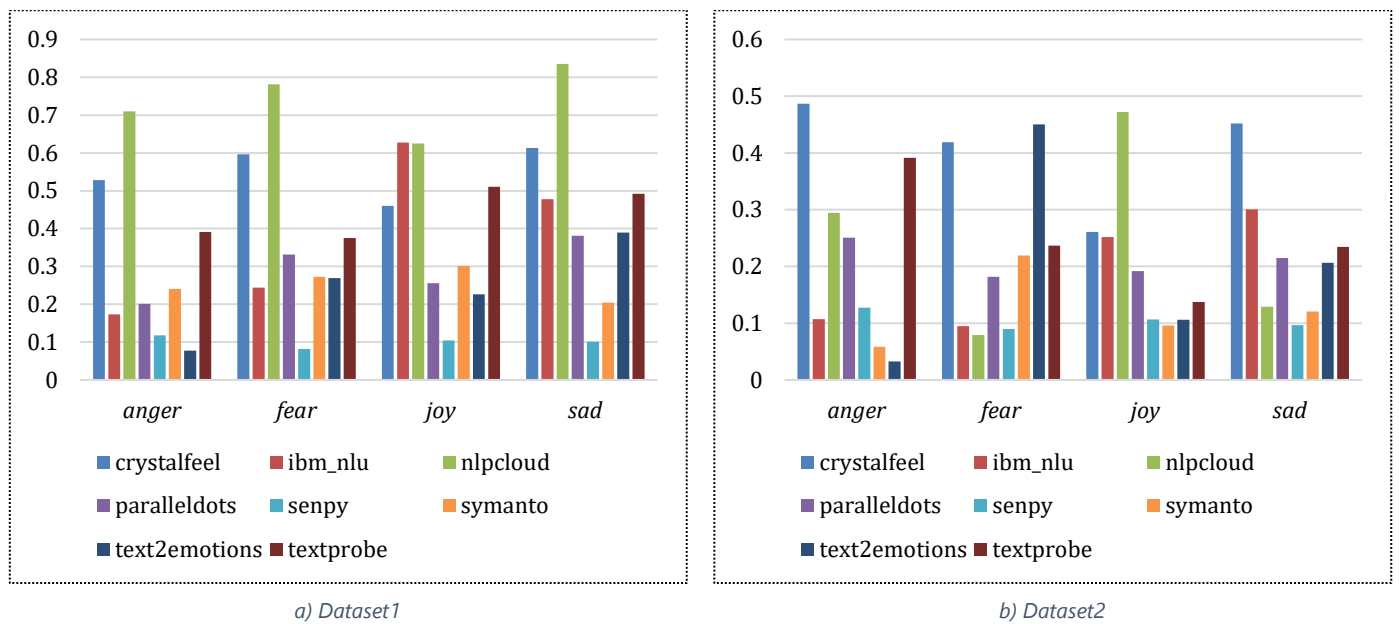

*a) Dataset1*    *b) Dataset2*

*Figure 10: The average of emotions likelihood values obtained by each API applied to each dataset.*

Figure 10 indicates an apparent variance between the extracted emotions of each model in each dataset. This can be observed in all emotions – despite some convergence between some APIs. This discrepancy between the extracted emotions and their likelihood values for each API emphasises the importance of conducting this study, thereby obtaining a better view of the reliability of these APIs to tackle the designated task. Nevertheless, the values depicted in Figure 10 verify the utility of Nlpcloud over dataset1 and Crystalfeel over dataset2.

## 5. Discussion

The tremendous technological development dominated by the emergence of online social networking sites has established an urgent need for automated processing of natural languages so as to understand and organise what is circulated on these sites (Abu-Salih et al. 2021b, 2021c, 2021a, 2021d). In fact, with the rise of social media, people started expressing their opinions more openly about their experiences with products and services through blogs, video blogs, social media stories, reviews, recommendations, hashtags, comments, replies, direct messages, news articles, and various other platforms. When this happens online, it leaves a digital fingerprint of an individual's expression of the experience. Now, this experience can convey one or more of various emotions, including *anger, joy, love, sadness,* and *fear*. Additionally, the widespread use of social networking sites has created a number of avenues for communication between businesses and their existing and potential clients. In fact, the accurate detection of emotions captured from the social textual contents represents a unique opportunity for companies to enhance the conversational dialogues between them and their customers. Therefore, developing advanced and sophisticated tools to detect textual emotions better is vital.

This study aims to furnish a comparative study between eight well-known APIs and technologies in terms of their effectiveness in detecting correct emotions from social textual contents. The study is conducted using two different datasets: (1) Emotions dataset for NLP and (2) Twitter Conversations dataset. Each API is used to extract the emotions from each dataset. Then correlation analysis was conducted on the resultant emotions, followed by undertaking a comparative study using various evaluation metrics. This study reveals specific as well as generic issues with the incorporated APIs that can be summarised as follows: to begin with, there is a clear divergence in both the type of emotion extracted by each API as well as the emotion likelihood values obtained by each API. This disagreement seems normal because each API uses their emotion detection algorithm(s), thus the results might differ. However, businesses and organisations that seek to extract emotions from their social media channels must scrutinise each API before selecting the best tool. This points to the significance of such studies that aim to help decision-makers by providing the necessary evaluation strategies (Al-Okaily, Alalwan, et al. 2022; Al-Okaily, Al-Okaily, et al. 2022; Al-Adwan et al. 2022; Al-Okaily, Al-Okaily, and Teoh 2021; Al-Qudah et al. 2022).

Additionally, this study emphasises the importance of conducting benchmark comparisons using unseen and untrained datasets. This is greatly important to ensure each API's fair performance outcome. For example, the experimental results demonstrate the superiority of Nlpcloud API in detecting emotions from dataset1 (Micro-F1 score = 0.99). However, the performance of this designated API notably degraded when applied to dataset2 (Micro F1 score = 0.55). Dataset1 is publicly available and is commonly used for NLP Classification and machine learning tasks. Hence, it is likely that this dataset has been used to train several emotion detection models including those incorporated in this study. On the other hand, this is the first study that uses dataset2, which is labelled mainly to measure the performance of the integrated APIs. Thus, the comparative results using this dataset offer a recent, objective, and unbiased evaluation.

This study fortifies efforts the aim of which is to provide a review of the current emotion detection APIs and tools. The embedded algorithms of these technologies are continuously evolving; thus, similar studies should be constantly conducted. This research also offers methodologies for academics and industrial practitioners to validate and verify the effectiveness of out-of-the-shelf emotion detection tools and APIs, especially with a lack of research to empirically compare current emotion detection technologies using social media datasets.

# 6. Conclusion

Advanced social media analytics is still being used by organizations to better understand the voice of their customers. In this paper, we provide a comparative study of using eight emotion-detection APIs on two datasets that were collected from social data. The included APIs were used to generate the emotions from the selected datasets. The aggregated scores provided by these APIs as well as theoretically supported assessment metrics such as the micro-average and macro-average of accuracy, classification error, precision, recall, and f1-score were used to evaluate the performance of these APIs. The evaluation of these APIs taking the evaluation measures into account is reported and discussed. The comparative study reveals important issues and emphesises the inadequacy of these APIs to handle new datasets. Therefore, there is need to continuously investigate novel techniques to furnish better emotion detection and recognition models. These models must attempt to tackle the sophisticated nature of this problem including multilingualism, ambiguity, colloquialisms, slang, irony, sarcasm, and contextual phrases and homonyms.

# Declarations

## CONFLICTS OF INTEREST

The authors declare no conflicts of interest.

## AUTHORS' CONTRIBUTIONS

Bilal Abu-Salih: Conceived and designed the experiments; Performed the experiments; Analyzed and interpreted the data; Contributed reagents, materials, analysis tools or data; and Wrote the paper.

Mohammad Alhabashneh: Analyzed and interpreted the data; Contributed reagents; and Wrote the paper.

Dengya Zhu: Contributed reagents, materials, analysis tools or data; and Wrote the paper.

Albara Awajan: Contributed reagents, materials, analysis tools or data; and Wrote the paper.

Yazan Alshamaileh: Analyzed and interpreted the data; and Wrote the paper.

Bashar Al-Shboul: Analyzed and interpreted the data; and Wrote the paper.

Mohammad Alshraideh: Analyzed and interpreted the data; and Wrote the paper.

## FUNDING STATEMENT

N/A

## ACKNOWLEDGMENTS

N/APage | 18